\documentclass[a4paper,11pt]{article}
\usepackage{graphicx}
\usepackage{algorithm2e}

\begin{document}

\title{Parallelization of the LBG Vector Quantization Algorithm for Shared Memory Systems}

\date{}
\author{\begin{tabular}[t]{c@{\extracolsep{2em}}c}
    Rajashekar Annaji \hspace{1in} Shrisha Rao \\
    {\tt \{annaji.rajashekar,srao\}@iiitb.ac.in} \\
    International Institute of Information Technology - Bangalore \\
    Bangalore 560 100 \\ India
\end{tabular}}

\maketitle

\begin{abstract}

  This paper proposes a parallel approach for the Vector Quantization
  (VQ) problem in image processing. VQ deals with codebook generation
  from the input training data set and replacement of any arbitrary
  data with the nearest codevector.  Most of the efforts in VQ have
  been directed towards designing parallel search algorithms for the
  codebook, and little has hitherto been done in evolving a
  parallelized procedure to obtain an optimum codebook.  This parallel
  algorithm addresses the problem of designing an optimum codebook
  using the traditional LBG type of vector quantization algorithm for
  shared memory systems and for the efficient usage of parallel
  processors.  Using the codebook formed from a training set, any
  arbitrary input data is replaced with the nearest codevector from
  the codebook.  The effectiveness of the proposed algorithm is
  indicated.

\end{abstract}

%\newpage

%\input{intro}

\section{Introduction}

Vector Quantization is an extension of the scalar quantization to
multi-dimentional space~\cite{S.Rungta}, which is a widely used
compresssion technique for speech and image coding
systems~\cite{troy}. It is similar to the clustering procedure known
as K-means algorithm~\cite{equitz} which is used in pattern
recognition. It is also useful in estimating the probability
distributions of the given featured vectors over a higher dimension
space.

%//may be this below para should come in BACKGROUND section // 
%The code vectors are distributed over the processors and the search necessary
%to coding an input vector is done in parallel. A parallel construction of the
%codebook using a distributed memory can be found.

For Data compression using Vector Quantization, the two main
distinguishable parts are codebook generation and replcement of
arbitrary data with the obtained codebook. This paper addresses the
parallel implementation of the codebook generation part.

Vector quantization involves the comparison of vectors in the input
training sequence with all the vectors in a codebook. Owing to the
fact that all the source vectors are compared with the same codebook
and the source (training) vectors are mutually exclusive in operation,
division of work can be clearly visualized and parallelization of the
algorithm can be achieved.
 
The LBG algorithm used for Vector Quantization needs to be modified by
exploiting the inherent parallelism, which should lead to the
improvement in the usage of processors. And this will inturn reduce
the time taken to generate the codebook for LBG algorithm.

This paper presents a parallel VQ algorithm based on the shared memory
architecture and uses the master/slave paradigm to optimize two main
bottlenecks of the sequential version:

\begin{itemize}
\item Time taken for the design of optimum codebook.
\item Efficient distribution of work to the parallel processing units.
\end{itemize}

Taking the advantage of shared memory architecture by assigning equal
chunks of input training data to all the processors and having a copy
of codebook in the primary memory of each processor, the `nearest
codevector identification' which is the most time consuming part in
the LBG algorithm is performed faster. The reduction in computation
time does not result in the increase in distortion. Distributing the
training samples over the local disks of slaves(processors) reduces
the overhead associated with the communication process.

The `cell table' which is formed after the integration by the master
is stored in shared memory and is used in the Updation procedure.
This is important for obtaining optimum codebook for the given input
training data.  Processors work parallelly, updating a single
codevector at any point. Assigning of the codevector to a processor is
done randomly.

The paper is organized as follows.  Section~\ref{related} describes
the related work of vector quantization in the field of data
compression and also provide some background about LBG and terms used
in the algorithm.  Section~\ref{System} describes the proposed Parallel
implementation of Vector Quantization
algorithm. Section~\ref{Performance} describes the performance issues
of the above proposed algorithm and speedup of the
system. Section~\ref{results} describes the results and simulations of
the algorithm using OpenMP.  Section~\ref{conclusion} concludes the
work done and describes some of the future work.

\section{Related Work} ~\label{related}

Considerable work is being done in the fields of image compression,
speech compression based on vector quantization and codebook
generation.  Some of the algorithms that have been used for sequential
VQ codebook generation are LBG, pair wise nearest neighbor(PNN),
simulated annealing, and fuzzy c-means clustering~\cite{equitz}
analysis. The compression method used for this is based on a block
encoding scheme known as vector
quantization~\cite{linde,Huang,equitz}.  The last decade has seen much
activity in exploring limits and applications for vector quantization.
Currently, parallel processing is being explored as a way to provide
computation speeds necessary for real time applications of image
compression techniques~\cite{troy}.
 
Codebook design plays a fundamental role in the performance of signal
compression systems based on VQ. The amount of compression is defined
in terms of the rate, which will be measured in bits per
sample. Suppose we have a codebook of size $N$, and the input vector
is of dimension $L$, in order to inform the decoder of which
code-vector was selected, we need to use $log_{2}N$ bits.  Thus the
number of bits per vector is $log_{2}N$. As each codevector contains
the reconstruction values for $L$ source output samples, the number of
bits per sample would be $\frac{log_{2}N}{L}$. Thus, the rate for an
$L$-dimensional VQ with a codebook size of $N$ is
$\frac{log_{2}N}{L}$.

The main problem related to VQ is that the training process to create
the codebook requires large computation time and memory, since at each
new experiment to evaluate new feature sets or increase in the
database for training the HMM's(Hidden Markov Models), it is necessary
to recreate the codebooks.  Parallelism operates on the fact that
large problems can almost always be divided into smaller ones, which
may be carried out concurrently. Based on this principle, an algorithm
for parallelizing of Vector Quantization (VQ) is proposed, which when
applied on a Shared Memory system like a Multicore system guarantees a
better and faster initialization method for LBG codebook design.
  
Codebook consist of a set of vectors, called codevectors. Vector
quantization algorithms use these codebook to map an input vector to
the codevector closest to it.  Data compression, the goal of vector
quantization, can then be achieved by transmitting or storing only the
index of the vector. Various algorithms have been developed to
generate codebooks. The most commonly known and used algorithm is the
Linde-Buzo-Gray (LBG) algorithm.

Parallel versions of the LBG Vector Quantization algorithm have been
proposed by many and most of them have been applied to Distributed
systems where the overhead of process communication is
present~\cite{troy,S.Rungta,H.J.Lee}. The speed and execution time of
the algorithm depends on these communications mechanisms, which are
minimal in case of shared memory architectures. Though some parts of
these parallel implementations are similar, the speed and
effectiveness depends upon the architecture used and efficient usage
of system processors and memories.

\subsection{Algorithms for Codebook Generation}

Various algorithms have been developed for codebook generation. Some
that have been used are the LBG, pair wise nearest neighbor (PNN),
simulated annealing and the fuzzy c-means (FCM) clustering algorithms.

\begin{enumerate}
    
\item LBG Algorithm~\cite{linde,Huang}: This is an iterative clustering
    descent algorithm. Its basic function is to minimize the distance between
    the training vector and the code vector that is closest to it.    
  
\item Pair wise Nearest Neighbor Algorithm~\cite{Akiyoshi}: This is
    a new alternative to the LBG algorithm and can be considered as an
    initializer for the LBG algorithm.  For efficient execution of
    this algorithm, the two closest training vectors have to be found
    and clustered. The two closest clusters are then merged to form
    one cluster and so on.
    
  \item Simulated Annealing Algorithm: This algorithm aims to find a
    globally optimum codebook as compared to the locally optimum
    codebook that is obtained using the LBG algorithm. Stochastic hill
    climbing is used to find a solution closer to the global minimum
    and escape the poor local minima.
    
  \item Fuzzy C-Means Clustering Algorithm\cite{equitz,Huang}: The LBG
    and PNN partition the training vectors into disjoint sets. The FCM
    algorithm assigns each training vector a membership function which
    indicates the degree to which it will belong to each cluster
    rather than assigning it to one cluster as is done in the LBG and
    PNN algorithms.

\end{enumerate}

Huang compares these algorithms~\cite{Huang}; the following are some of his
conclusions.

\begin{itemize}

\item PNN is the fastest but has higher distortion than the LBG
  algorithm~\cite{Akiyoshi}.
    
\item Simulated Annealing produces the best distortion results but
  requires substantially greater CPU time and there is no significant
  improvement in the quality of the reproduced images.
    
\item The FCM algorithm has worse results and is slower than the LBG
  algorithm.

\end{itemize}

\subsection{Background}

Looking at the different codebook generations and their drawbacks, the
algorithm chosen for parallelizing was the LBG algorithm. An attempt
has been made to decrease the time for codebook generation while
maintaining the distortion measures obtained from the sequential
version of the LBG algorithm. The next section describes the LBG
algorithm in more detail which is taken from the original Vector
Quantization paper~\cite{linde}.

\textbf{LBG Algorithm}
\begin{itemize}
\item 
Initialization: Given $N$ = number of levels, distortion threshold \( \epsilon \geq 0 \), an initial $N$-level reproduction alphabet \( A_0 \) and a training sequence \(\{x_j,;j=0...n-1 \}\) . Set \( m=0 \) and \( D_{-1} = \infty \).
\item
Given \(A_m = \{ y_i;i=1....N\} \) find the minimum distortion partition 
\(P (A_m)=\{S_i;i=1...N\} \) of the training sequence: \( x_j \in S_i \) if \( d(x_j,y_i) \leq d(x_j,y_l)\), for all $l$. Compute the average distortion 
\( D_m = D(\{A_m,P(A_m)\})=n^{-1}\sum^{n-1}_{j=0} min_{y\in A_{m}} d(x_j,y).\)
\item
If \(\frac{(D_{m-1}-D_m)}{D_m} \leq \epsilon \), halt with \(A_m\) final reproduction alphabet. Otherwise continue.
\item
Find the optimal reproduction alphabet \(x(P(A_m))=\{x(S_i);i=1,..,N\}\) for \( P(A_m)\). Set \(A_{m+1} \equiv x(P(A_m))\). Replace $m$ by $m+1$ and go to $1$. 

\end{itemize}

%The main drawbacks of sequential LBG algorithm is the time it takes to
%create the codebook and its updation to produce the local maximum
%codebook for the training set.

\noindent\textbf{Terms used:}

\begin{itemize}
  
\item \textbf{Codebook:} It is a collection of codevectors,and these
  codevectors can be stated as the quantization levels. In LBG
  algorithm number of codevectors can be only in powers of $2$.
  
\item \textbf{Centroid:} Centroid is nothing but the average of the
  input vectors in the particular region specified. The dimension of
  the centroid is same as Input training vector $k$.
  
\item \textbf{Partial Cell Table:} It is the part of the 'cell table'
  and it indicates the allocation of input vectors to the corresponing
  minimum distortion codebook. Against each index it stores the
  codevector number from which the corresponding input vector has the
  minimum distance(eucledian distance). Each processor has its own
  partial cell table.
  
\item \textbf{Cell Table:} After all the processors compute the
  minimum distortions and partial cell tables are formed, they are
  integrated to form the final 'cell table' and the values in the
  table are called as cell value.
  
\item \textbf{Distortion:} The Eucledian distance between the input
  training vector and the codebook vector gives the distortion
  value. It helps in identifying the nearest codebook.

\end{itemize}

\section{System Model for Parallel implementation of Vector Quantization} ~\label{System}

This parallel algorithm can be extended to shared memory architectures
where in, all the processor cores have their primary cache memory,
also called as $L1$ cache.  And a single shared memory also called as
$L2$ cache, which is accesible by all the cores by some means
interprocess communication.  The input training data is available in
the shared memory and hence can be shared by all the processors.

\subsection{Notations}

We assume that the initial training data which is used to train the
system is of the form of an $M \times k$ matrix where $k$ is the vector
dimension.  This can be any numerical data, which is obtained by
processing an image or a speech signal.

\begin{itemize} 

    \item Number of Input Training vectors : $M$
    \item The dimension of Input data: $1 \times k$
    \item Codebook size: $N \times k$, where $N$ is the number of
      codevectors
    \item Number of processors       : $P$

\end{itemize}

The set of training vectors is
%\begin{center}
\[\tau = \{X_{1},\ldots,X_{M}\}\] 
%\end{center}
and each input vector is denoted by,
%\begin{center}
\[X_{m}=\{x_{m1},\ldots,x_{mk}\}, m=1,2,\ldots,M\]
%\end{center}
%'$N$' is the number of codevectors, hence
%\begin{center}
\[\mathbf{C}=\{C_{1},\ldots,C_{N}\}\]
%\end{center}
represents the codebook.  Each codevector is $k$-dimensional, e.g.,
%\begin{center}
\[C_{n}=\{c_{n1},\ldots,c_{nk}\}, n=1,2,\ldots,N\]
%\end{center}

\subsection{Algorithm Implementation}
%(The pseudo code must be here)\\

%\input{NewCode}

%\incmargin{1em}
%\restylealgo{boxed}
\linesnumbered
\begin{algorithm}
  
  \SetKwData{Left}{left}
  \SetKwData{This}{this}
  \SetKwData{Up}{up}
  \SetKwData{MinDist}{MinDist}
  \SetKwData{index}{index}
  
  \SetKwFunction{Union}{Union}
  \SetKwFunction{FindCompress}{FindCompress}
  \SetKwFunction{MasterSelection}{MasterSelection}
  \SetKwFunction{PartialCellTable}{PartialCellTable}
  \SetKwFunction{minimum}{minimum}
  \SetKwFunction{centroid}{centroid}
  \SetKwFunction{CodebookIndex}{CodebookIndex}\Indm
  \SetKwFunction{CommunicateToMaster}{CommunicateToMaster} 
  \SetKwFunction{IntegratePartialCellTables}{IntegratePartialCellTables}
  \SetKwFunction{celltable}{celltable}
  \SetKwFunction{IntegrateDistortions}{IntegrateDistortions}
  \SetKwFunction{NewCodevector}{NewCodevector} 
  \SetKwFunction{ExtractVectors}{ExtractVectors}
  
  \SetKwInOut{Input}{input}
  \SetKwInOut{Output}{output}
  
  \caption{Parallelized Vector Quantization Algorithm}
  \Input{any data of dimension $M \times k$}
  \Output{codebook of dimension $N \times k$}
  \BlankLine
%  \emph{special treatment of the first line}\;
\Indp
\emph{\MasterSelection{}}\;
 \Indp
  $\rightarrow$ compute \centroid(input)\;
  $\rightarrow$ \emph{Distribute inputs to Processors}
  \BlankLine
 \Indm
  \textbf{\emph{Codebook Splitting:}}\\ 
   \Indp \lForEach{$centroid$}{
    \emph{[centroid + $\delta$],[centroid - $\delta$]}
  \BlankLine
   \Indm
  \textbf{\emph{Cell Allocation:}}\\
  \Indp  
  $\rightarrow$ \emph{parallel execution at each processor with processor ID
  $\rho$ = ${0,1,\ldots,P-1}$}\;
  \{\\
   \Indp
   \For{$z\leftarrow [\rho \times \frac{M}{P}+1]$ \KwTo
    $[(\rho+1) \times \frac{M}{P}]$}{\nllabel{forins}
    \MinDist $\leftarrow$ \minimum{$E.Dist[input(z),codebook(0)],E.Dist[input(z),codebook(1)]\ldots $}\; \index $\leftarrow$ \CodebookIndex{$MinDist$}\; $D_{\rho}$ $\leftarrow$ $D_{\rho}$ + \MinDist \; 
    \PartialCellTable{$z$}$\leftarrow$ \index\; }
   \Indm
   \}\\
    \Indm
    \BlankLine
    \CommunicateToMaster{}\;
    \celltable{} $\leftarrow$ \IntegratePartialCellTables{}\;
    $TD_{\rho}$ $\leftarrow$ \IntegrateDistortions{}\;
    \BlankLine
      \If{$\frac{TD_{\rho} - TD_{\rho + 1}}{TD_{\rho}}$ $\leq$
      $\epsilon$(Threshold)} { 
      		\lIf{$Vectors$ $==$ $N$(required number of codevectors)}
      			{$TERMINATE$\;}
        	\lElse{go to \emph{CodebookSplitting} step}\\
    	}
    \Indm	
    \Indp
    \textbf{else}
    \emph{\textbf{Updation step:}}\\
    \Indp
 	\emph{$\rightarrow$ parallel execution at each processor}\;
 	\{\\
 	\Indp
	\For{$j \leftarrow 1 $ \KwTo $M$}{\nllabel{forins}
	 	\ExtractVectors{$CellTable,index$}\;
	 	\NewCodevector{$index$} $\leftarrow$ \centroid{$ExtractedVectors$}\;
    	}
    \Indm	
    \}\\	
    	go to \emph{Cell Allocation} step\; 
    }
 
   \label{Parallel Vector Qunatization algorithm}
\end{algorithm}
\decmargin{1em}

The parallel algorithm is given as follows.

Functions used : \\
\emph{Centroid(Input)} $\rightarrow$ The average of input.\\
\emph{CodebookIndex(MinDist)} $\rightarrow$ This function gives the index of
	the codevector to which the input is nearest and the argument is 'minimum
	euclidian distance'.\\
\emph{CommunicateToMaster()} $\rightarrow$ All the processors communicate to
	the master to integrate the parallel sections.\\
\emph{IntegratePartialCellTables()} $\rightarrow$ The partial cell tables
	formed by individual processors are integrated to form the final 'cell
	table'.\\ 
\emph{IntegrateDistortions()} $\rightarrow$ The distortion value obtained by
	each processor for its set of input vectors allocated is communicated to the
	master, which does the integration of all those distortions.\\
\emph{ExtractVectors(celltable,index)} $\rightarrow$ The values in the
	celltable are the indexes of codevectors. From the 'CellTable' extract all the
	input vectors which belong to a particular codevector denoted by its index.\\
\emph{NewCodevector(extracted vectors)} $\rightarrow$ The centroid of the
  	extracted vectors gives the updated codevector.\\\\
The following are the important steps in the Algorithm:
\begin{enumerate}
  \item Initialization
  \item Codebook Splitting
  \item Cell Allocation
  \item Updation
\end{enumerate}

\noindent\textbf{Initialization:}

\begin{itemize}
\item One of the processors is chosen as the master either by leader
election or randomly.
\item The master computes the centroid of the initial
training data and it is assigned as the initial codebook .

\begin{table}[!ht]
\centering
\begin{tabular}{|c|c|c|c|c|}
    \hline
    \multicolumn{5}{|c|}{Input Training set} \\
    \hline
    1 & $a_{11}$ & $a_{12}$ & .... & $a_{1k}$ \\
    \hline
    2 & $a_{21}$ & $a_{22}$ &  & $a_{2k}$ \\
    \hline
    : &  &  &  &  \\
    \hline
    : &  &  &  &  \\
    \hline
    M & $a_{m1}$ & $a_{m2}$ & .... & $a_{mk}$ \\
    \hline
    & \multicolumn{4}{c|}{Centroid} \\
    \hline
    \end{tabular}
\caption{Centroid}
\end{table} 

In Table 1, the centroid ($1 \times k$) is the average of all the
input training vectors of the dimension $M \times k$, and forms the
initial codebook.

\item The master allocates the training vectors equally among all the
  slaves. The number of vectors to each slave are
  \(\lfloor\frac{M}{P}\rfloor\).
    
\item $D_{-1}$ which is the distortion value, is initialized to a very
  high positive value.
    
\item The threshold $\epsilon$ decides the termination condition of
  the algorithm.
    
\item The splitting parameter '$\delta$' is initialized to a constant value
    .
\end{itemize}
%\Indm

\noindent\textbf{Codebook Splitting:}

The master doubles the size of the initial codebook.  The increase in
the number of codevectors is done by getting two new values, by adding
and subtracting $\delta$ (which may be considered a variance, and is
constant throughout the algorithm), to each centroid value.
Therefore, the codebook splitting step generates a new codebook which
is twice the size of the initial codebook.
 	
%\begin{center}
\[[Centroid + \delta], [Centroid - \delta]\]
%\end{center}

The codebook so formed in this splitting step is duplicated into the
primary memory of all the slave processors.

\noindent\textbf{Cell Allocation:} \\

Each slave calculates the Euclidian distance between the input vectors
allocated to it and the codebook vectors.  Based on the Euclidian
distance, it finds the nearest codebook vector to each input vector
and allocates the particular input vector to the codebook vector.

\begin{figure}[!ht]
\centering
	\includegraphics[height=3.0in]{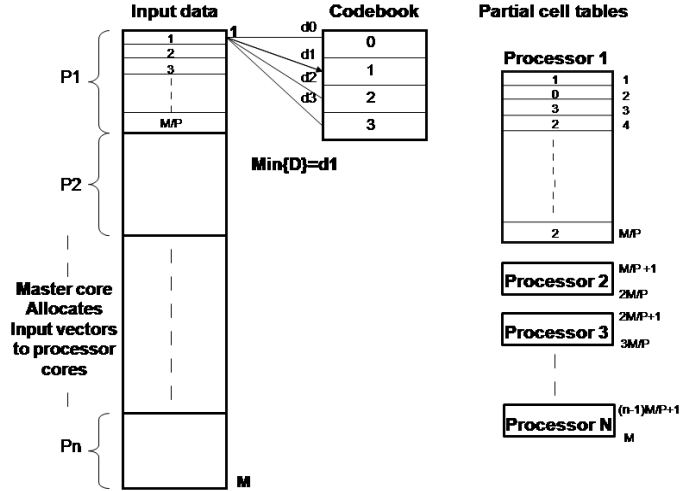} 
\caption{Input data allocation to processors and nearest codebook , cell table}
\end{figure}

In Figure 1, the initial training data is divided equally among all
the processors $\mathbf{P}=\{P_{1} \ldots P_{n}\}$ . $processor1
\leftarrow$ [$1$ to $\frac{M}{P}$], $processor2 \leftarrow$
[$(\frac{M}{P}+1)$ to $\frac{2M}{P}$], $\ldots$ $processorN
\leftarrow$ [$((n-1)\frac{M}{P} +1)$ to $M$].  Whenever a new codebook
is formed, codevectors are duplicated into primary cache($L1$) of all
slave processors.  The Euclidian distance between each input vector
allocated to the processor and all the codevectors is computed,
minimum of all is taken and added to '$D_{P_{i}}(Distortion)$'. And
the index of the codevector nearest to the input is placed in the
corresponding location in the `partial cell table' ($\frac{M}{P}
\times k$).  Finally when all the allocated training vectors are
computed, the `partial cell table' and `distortion' have to be
communicated to the master, and this process is done by every other
processor executing in parallel.

For each slave processor, the distortion value is
$\mathbf{D}_{i}$=$\Sigma min\{$D$\}$ where $D=\{d_{1},\ldots,d_{r}\}$
is the set of Euclidian distortions for each input vector with respect
to the codevectors.  And the corresponding `distortion' values and
`partial cell tables' will be communicated to the master by all the
slaves, and the master computes the `total distortion' $TD_{i}$ and
also integrates the individual cell table to form a `final cell table'
which is used for updating codebook.

The total distortion computed by the master is
%\begin{center}
\[TD_{i} = \Sigma \{D_{i}\}\]
%\end{center}

The threshold value $\epsilon$ which is initialized previously to a
constant value is used to identify the termination of the
algorithm. The termination condition for the algorithm is,
%\begin{center}
\[\frac{( TD_{i-1}-TD_{i} )}{(TD_{i-1})} \leq \epsilon\]
%\end{center}

If the condition is satisfied, it implies that the optimum codebook
for that level is formed for the given input training data set.  And
if the number of codevectors formed in that state is equal to the size
of the codebook specified, terminate the program.  Else go back to the
Codebook Splitting step and proceed with the same.

\begin{figure}[!ht]
\centering
	\includegraphics[height=2.5in]{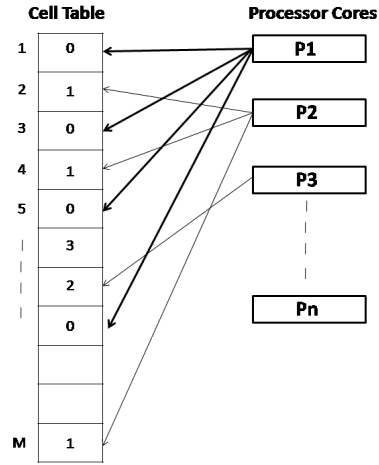} 
\caption{Parallel updation of codebook by slave processors}
\end{figure}

\noindent\textbf{Codebook Updation:}\\

If the Threshold condition is not satisfied then it implies that the
codebook so formed is not the optimum one and needs to be updated. The
codebook is updated by replacing each codevector by the centroid of
all the input vectors allocated to it.  The new centroids are
calculated parallely, with set of input vectors corresponding to one
codevector are computed by a single processor.  This procedure is
explained from Figure 2.

In Figure 2, the cell table contains the indexes of codevectors to
which the corresponding input vector is nearest. A single processor
updates single codevector at any point.  The process of updation is:

\begin{itemize}
  
\item Extract all the input vectors which have the value $0$ in the
  cell table. These are the set of input vectors which are nearest to
  codevector1.
  
\item Compute the centroid of the vectors which have been
  extracted.  This forms the new codevector.

\end{itemize}

Hence, if we have $P$ slave processors available, then at any point of
time $P$ codevectors will be updated in parallel, and then the
next round of updation proceeds and so on upto the number of
codevectors, which executes in a round robin fashion.  And the
assigning of $P$ codevectors to the $P$ processors can be done
randomly or serially. In the case of serial updation, if the size of
the codebook in that stage is '$S$', then $S$ number of codevectors
must be updated.  If $CodevectorIndex$ \% $P$ == $\phi$, implies
processor $\phi$ performs the updation of that codevector. Once all
the codevectors are updated go back to Cell Allocation step and
continue the iterative procedure until the required number of
codevectors are generated.

\section{Performance} ~\label{Performance}

From Algorithm 1 described in the Section~\ref{System} it can be
observed that the parallel processing tasks are identified
separately. And from our experimental analysis, in the sequential
version of Vector quantization using LBG algorithm, these parts of the
program which can be parallelized were extracted out as separate
process blocks (functions). And using a GNU {\tt gprof} profiler, which
helps in knowing where the program has spent most of its time and
which process is called by which other process while it is executing,
it is observed that parallel blocks consume $80--85$\% of the
overall time and also provides the time consumed by individual
parallel blocks.  This parallelization can be accounted to two of the
main processes in the algorithm.

\begin{enumerate}

\item Generation of `Cell Table' also called as `Allocation table,'
  i.e., allocating the input training vector to the nearest codebook
  vector based on Eucledian distance, which is a highly time consuming
  comuputation.

\item Calculation of Centroids in the Updation of codebook step.

\end{enumerate}

The first step is the most time-consuming part and takes about
$70--75$ \% time of the total sequential part. In this step the entire
input training data is divided into number of chunks equal to number
of processors and allocated to them.  As each processor has its own
duplicated version of the codebook, so further computations are done
in parallel by the processors.  Hence, the theoretical efficiency of the
multicore system or to say processor utilization' would be 100\%.
The second step, which is calculation of centroids in the updation
step, takes about $10--15$ \% of the sequential part.  According to
Amdahl's law:

\begin{center}
speedup=$\frac{1}{(S+\frac{(1-S)}{n})}$ 
\end{center}

where $S$ is time spent in executing the serial part of the
parallelized version and $n$ is the number of parallel processors.  In
the proposed algorithm , $85$\% of it can be parallelized, hence the
time spent in executing serial part is $15$\% and assuming $n$ =4, the
speedup of the parallel version is $1/(0.15+(1-0.15)/4) = 2.76$, i.e.,
a quadcore system with has $4$ cores would work $2.76$ times as fast
as the single processor system.

\section{Results and Simulation}~\label{results}

The proposed algorithm has been simulated using OpenMP with the
training input size of $2000$ vectors with the dimension varying from
$2$ to $10$.

These results have been compared with the standard sequential
execution and the results have been plotted. The graph clearly
indicates the expected reduction in time consumed to create the
codebooks.

\begin{figure}[!ht]
\centering
	\includegraphics[height=2.5in]{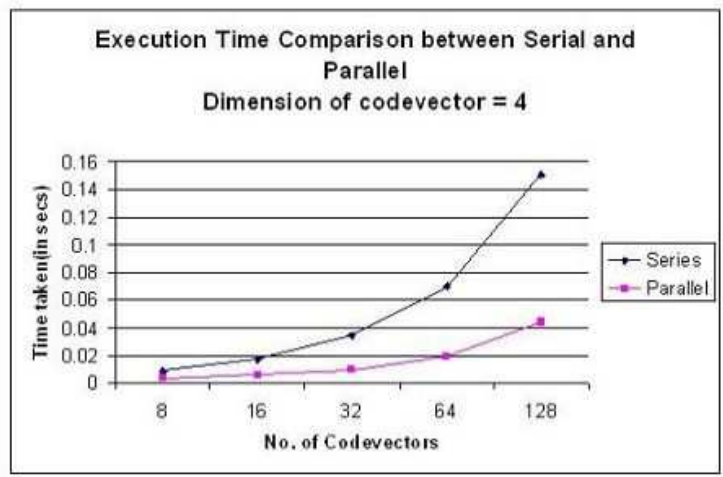} 
\caption{Execution Time Comparision}
\end{figure}

Figure 3 is the graph of the number of codevectors vs time taken to
create the codebook. The number of processors are fixed at $4$ and the
number of code vectors are varied from $8$ to $128$.  The results
are plotted both for sequential as well as parallel algorithm.  The
difference is clear when the size of the codebook increases.
 
\begin{figure}[!ht]
\centering
	\includegraphics[height=2.5in]{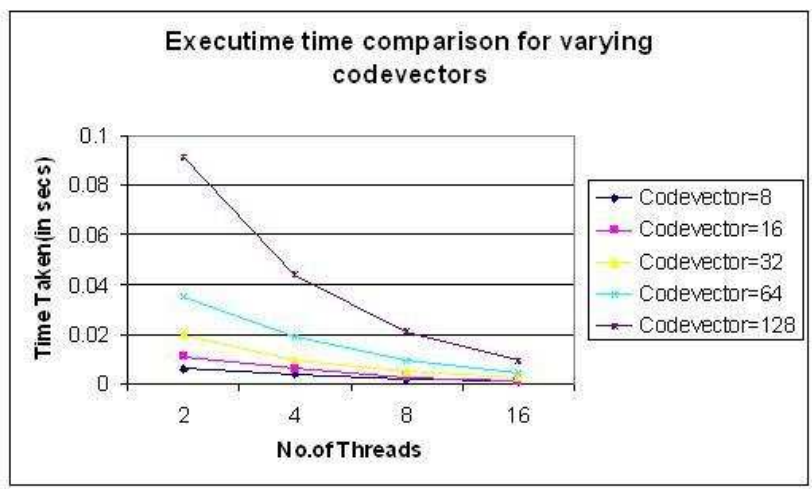} 
\caption{Plot of Time Taken vs. Number of threads}
\end{figure}

Execution time of the parallel algorithm is not exactly half of the
sequential time for a dual-core.

%\input{conclusion}
%\newpage

\section{Conclusions and Future Work} ~\label{conclusion}

In this paper a parallel implementation of LBG Vector quantization is
demonstrated.  The algorithm proposed here is general in the sense
that it can be applied on any shared memory architecture irrespective
of the underlying interprocess communication.  The performance of the
proposed algorithm is analyzed and also experimentally simulated using
the OpenMP shared programming.  When compared with the sequential
version of the LBG, the proposed algoritm has better performance in
terms of speed of execution, and the study of execution time is done
for varying number of parallel processing units.

The results obtained are the output of simulations of the sequential
version of VQ using OpenMP.  Implementation of the same using a
multicore system would provide accurate results regarding aspects like
`memory used' and `core utilization' which were difficult to obtain in
the present scenario.  In a typical low-bandwidth network, consider a
server which provides videos on demand.  If the number of requests to
the server is very high then the server sends a compressed version of
the original high quality video.  In this kind of scenario, the
algorithm proposed here would greatly enhance the server response time
by decreasing the time taken for video compression techniques which
make use of lbg algorithm.

\end{document}